\documentclass[letterpaper, 10 pt, conference]{ieeeconf}  

\IEEEoverridecommandlockouts                              

\usepackage[x11names]{xcolor}

\usepackage{soul}

\usepackage{graphicx}

\def\shrug{\texttt{\raisebox{0.75em}{\char`\_}\char`\\\char`\_\kern-0.5ex(\kern-0.25ex\raisebox{0.25ex}{\rotatebox{45}{\raisebox{-.75ex}"\kern-1.5ex\rotatebox{-90})}}\kern-0.5ex)\kern-0.5ex\char`\_/\raisebox{0.75em}{\char`\_}}}

\newcommand{\shruggies}[1]{
\tiny\shrug
\scriptsize\shrug
\footnotesize\shrug
\small\shrug
\normalsize\shrug
\large\shrug
\Large\shrug
\huge\shrug
\Huge\shrug
}

\newcommand{\shruggiesCol}[1]{
\tiny\shrug\\
\scriptsize\shrug\\
\footnotesize\shrug\\
\small\shrug\\
\normalsize\shrug\\
\large\shrug\\
\Large\shrug\\
\huge\shrug\\
\Huge\shrug\\
}

\newcommand\tab[1][5mm]{\hspace*{#1}}

\newcommand{\citethis}[1]{\textcolor{red}{[CITE]} }

\newcommand{\inst}[1]{$^ {#1} $}

\usepackage{amsmath}
\usepackage{amsfonts}
\usepackage{url} 
\usepackage[long]{optidef}
\usepackage{mathtools}
\usepackage{ragged2e}
\usepackage{tabularx}
\usepackage{lipsum}
\usepackage[noadjust]{cite}

\begin{document}
\pagenumbering{Roman}

\title{\LARGE \bf
Cooperative Modular Manipulation with Numerous Cable-Driven Robots for Assistive Construction and Gap Crossing }

\author{Kevin Murphy\inst{1,2} \and Joao C. V. Soares\inst{1,3} \and Justin K. Yim\inst{1} \and Dustin Nottage\inst{2} \and Ahmet Soylemezoglu\inst{2} \and  Joao Ramos\inst{1}
\thanks{*This work was supported by the US Army Corps of Engineers Engineering Research and Development Center via award number W9132T22C0013}
\thanks{$^1$Department of Mechanical Science and Engineering at the University of Illinois at Urbana-Champaign, USA}
\thanks{$^2$ US Army Corps of Engineers Construction Engineering Research Laboratory, Champaign, IL 61822, USA}
\thanks{$^3$ Dynamic Legged Systems Laboratory, Istituto Italiano di Tecnologia (IIT), Genova, Italy}
}

\maketitle
\thispagestyle{empty}
\pagestyle{empty}

%
\begin{abstract}\\
Soldiers in the field often need to cross negative obstacles, such as rivers or canyons, to reach goals or safety. Military gap crossing involves on-site temporary bridges construction. However, this procedure is conducted with dangerous, time and labor intensive operations, and specialized machinery. We envision a scalable robotic solution inspired by advancements in force-controlled and Cable Driven Parallel Robots (CDPRs); this solution can address the challenges inherent in this transportation problem, achieving fast, efficient, and safe deployment and field operations. We introduce the embodied vision in Co3MaNDR, a solution to the military gap crossing problem, a distributed robot consisting of several modules simultaneously pulling on a central payload, controlling the cables' tensions  to achieve complex objectives, such as precise trajectory tracking or force amplification. Hardware experiments demonstrate teleoperation of a payload, trajectory following, and the sensing and amplification of operators' applied physical forces during slow operations. An operator was shown to manipulate a 27.2 kg (60 lb) payload with an average force utilization of 14.5\% of its weight. Results indicate that the system can be scaled up to heavier payloads without compromising performance or introducing superfluous complexity. This research lays a foundation to expand CDPR technology to uncoordinated and unstable mobile platforms in unknown environments. 


\vspace{-1mm}
\end{abstract}

\section{Introduction}
Military gap crossing involves on-site construction of temporary bridges to overcome negative obstacles and increase field unit mobility and safety. Emplacing bridges allows rapid, efficient deployment and maintenance of supply lines supporting the movement of troops and equipment during combat operations. While beneficial, military bridging comprises several formidable challenges~\cite{army_bridging,bridging_army_doctrine}. 


Current methods require assembly of bridging components using specialized vehicles and heavy equipment. This introduces complex logistics for acquiring, transporting and deploying the necessary equipment and specially trained operators. Using copious time and hazardous manual labor, construction proceeds section by section on rollers, which facilitate movement into final locations over obstacle areas~\cite{mabey_bridge_compact_200}. Harsh field conditions such as swamps, aquatic areas, or austere terrains, add challenges for heavy machinery and personnel. Danger is only increased in contested zones.

Future military bridging requires a system with the flexibility to manipulate diverse bridges and components, operate in close proximity to humans, and provide a large workspace for effective operations. Meeting these capabilities will enhance the efficiency and effectiveness of field gap crossing, ensuring safety, and boost flexibility and capabilities for field units in diverse operational environments.

\begin{center}
\begin{figure}[t]
    \centering
    \includegraphics[width=0.485\textwidth]{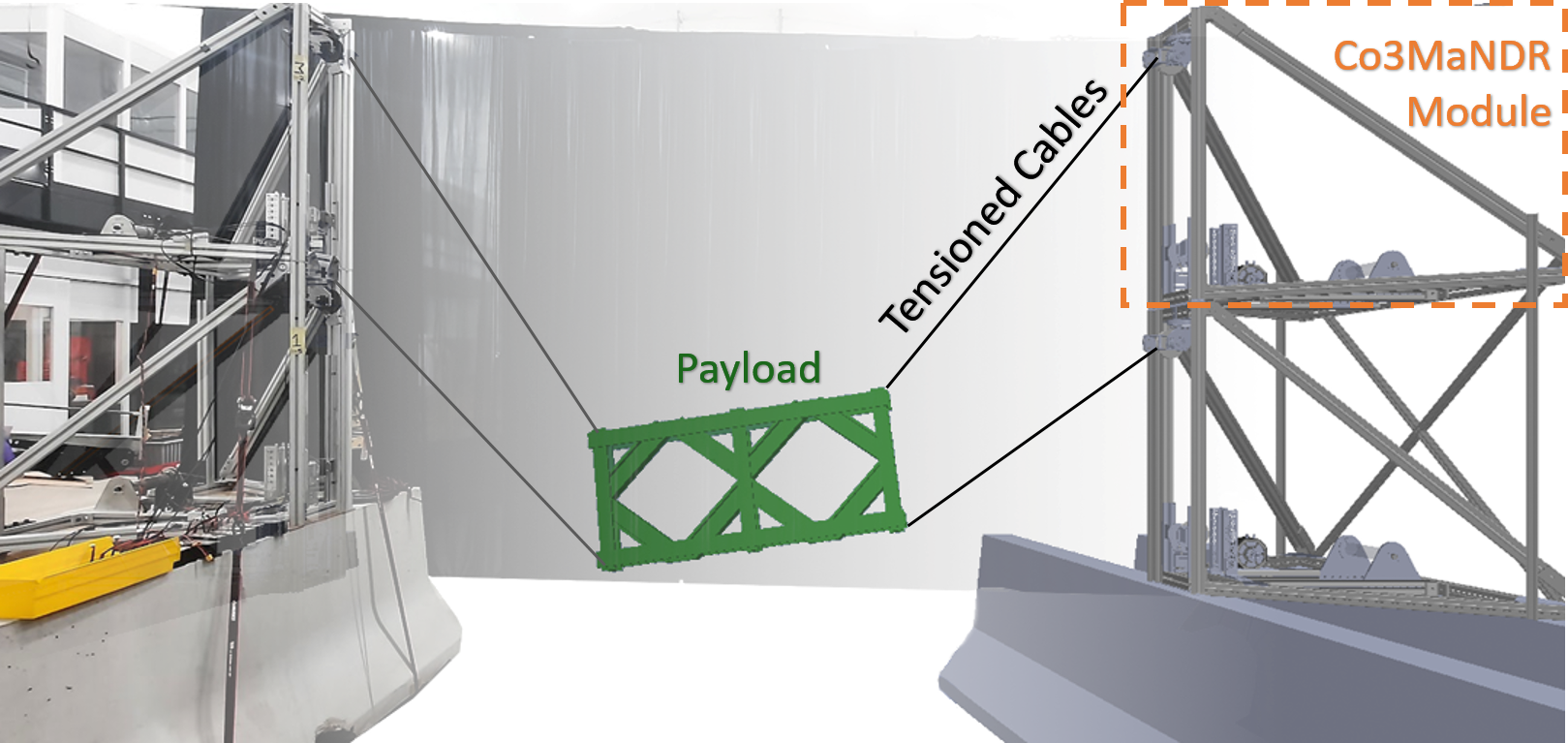}
\caption{Co3MaNDR System, consisting of four modules (orange box) in planar configuration. The experimental setup (left) and CAD model (right) are shown overlayed onto each other, with an example payload (center).  } 
\label{fig:entire_system}
\vspace{-7mm}
\end{figure}
\end{center}

This project addresses these challenges and pioneers a viable method which enables a lone person to install gap crossing solutions without heavy equipment. Our research introduces a modular robotic manipulation system (Fig. \ref{fig:entire_system}), where multiple robots independently act as force sources on a central payload, providing versatile solutions for large payload manipulation bridging, civil, and industrial sectors.

\vspace{-10mm}

\subsection{Related Works}


Co3MaNDR integrates three key elements in its framework: the innovative use of Cable Driven Parallel Robots (CDPRs) through a legged locomotion perspective, force control, and optimization techniques.


\subsubsection{Force Control}
In contrast to position or velocity control, force control focuses on regulating forces, which is important during interactions with unknown objects/environments. This has been demonstrated by several robots, especially in tasks involving manipulation and locomotion, through the adoption of force-controlled designs and strategies. Notable examples include Boston Dyanamics' Spot \cite{BD_spot} and Atlas \cite{BD_atlas}, the MIT mini cheetah \cite{cheetah}, and LIMS2-Ambidex \cite{limsII}. These robots' successes stem from the adoption of specialized hardware. There are two main types of electric actuators used for force control. Proprioceptive Actuators and Series Elastic Actuators (SEAs) have both been shown to be capable of force control and have their individual advantages and disadvantages~\cite{proprio,cheetah,SEA}. This project aims to support dynamic motions that require high mechanical bandwidth, leading to the use of proprioceptive actuators.




\subsubsection{Cable-Driven Parallel Robots}

CDPRs are a class of robotic systems characterized by controlling an end-effector with parallel cable-based mechanisms. CDPRs generally consist of stationary winches surrounding the workspace connected to the end-effector with cables. CDPRs do not move serial components or the motors themselves; therefore, resulting inertia is due to only the payload and reflected actuator inertia. CDPRs are lightweight and have high payload-to-mass ratios compared to traditional robotic manipulators~\cite{CDPR_survey}. While CDPRs are known for their adaptability and reconfigurability, they experience unique control challenges such as cable intersections limiting rotations, cables' inability to push, cable stretch, and sag \cite{sag}. CDPR have been seen in a vast array of applications, including flight simulation~\cite{helicopter}, construction and cinema~\cite{CDPR_survey}, logistics~\cite{FASTKIT}, marine platforms~\cite{HOROUB2018314}, painting~\cite{SERIANI20151}, rehabilitation~\cite{ABBASNEJAD20161}, motion simulation~\cite{SCHENK2018161}, and additive manufacturing~\cite{3D_print}.

The characteristics of CDPRs are well-suited to meet the demands of the project. CDPR’s actuators distributed around the workspace suggests distributing the robots as modular payloads onto several platforms or the environment. Module placement is a design choice; the system can easily reconfigure the placement and interaction of modules to match situational requirements. Similarly, scaling the number of modules allows scaling total system lifting capabilities, making the system payload agnostic. Finally, CDPRs customarily cover expansive workspaces while maintaining dexterity and accuracy, which aligns with construction requirements. 

However, traditional CDPRs are not well suited to field construction. CDPRs generally use position control on a system with highly geared actuators, which preclude high bandwidth force control. Second, CDPRs are usually based off external rigid framing surrounding the workspace~\cite{CDPR_survey, ABBASNEJAD20161, SCHENK2018161, 3D_print, sag, closed_loop_force_control, wrench_feasible_workspace}. The necessity for this additional structure undermines the objective of the proposed work. Closed-loop force feedback has been shown in CDPRs \cite{closed_loop_force_control}, but the hardware and control approach often prevents rapidly regulating forces, which are necessary for interactions with human and unknown environments.

\subsubsection{Multi-Agent Manipulation}
Multiple agents connected to the same rigid body creates closed kinematic chains, greatly increasing the complexity of the control problem. Frictional effects increase non-linearly and rotor inertia is amplified by the square of the gear ratio~\cite{youngwoo , youngwoo_b}, making traditional actuators with large gear ratios non-backdrivable. In such systems, small pose errors cause substantial interaction forces which can damage the robots and the payload. This makes Multi-Agent Manipulation exceptionally challenging for traditional robotics. 

Heavily controlled motions have been achieved~\cite{bimanual}, but extremely slowly in disturbance free, known, and controlled environments. Manipulating deformable payloads~\cite{deform_a, deform_b} circumvents issues with kinematic error. However, the specialized flexible payload are not applicable to construction or structures. Similarly, field operations incur speed requirements and slow operations. Even if they could be achieved in unknown and uncontrolled environments, personnel's presence in hazardous conditions would be extended. 

An exciting project seen in \cite{Lynch} shows an SEA and omni-wheel based distributed robotic system that also shows human guided force amplification. However, this system is limited by the required controlled flat ground, the mobile module's footprint, and rigid parallel linkage connections reducing the range of motion of the payload.

\subsection{Approach and Contributions}
We propose field bridging and cumbersome payload manipulation can be achieved with Cooperative Mobile and Modular Manipulation with Numerous cable-Driven Robots (Co3MaNDR), a system engineered to exhibit critical system attributes: modular, physically reconfigurable, compliant, backdrivable, and force controlled. 

Co3MaNDR's approach shifts several paradigms, which informs system level design, dictates actuator and sensor selection, and structures the chosen control method. First, this approach replaces conventional employment of specialized heavy equipment with distributed, smaller collaborative modular robots. Co3MaNDR adopts a payload-centric controller based on interaction forces, deviating from the traditional position control methods employed in CDPRs and allowing system level force sensing and motions guided by operators physically applying force. Individual modules plan together as a single unit, but can act independently without coordination, increasing agility and adaptability in dynamic field environments. The variable number of modules makes optimization-based control methods particularly applicable. This innovative approach leverages the inherent advantages of distributed modular robotics and aligns seamlessly with the CDPR-based system design to introduce new capabilities.

We designed and implemented this novel scalable system design, demonstrating the same operations with two and four modular robots. In these experiments, four robots exerted on average half the torque compared to two robots. This suggests that heavier payloads can be manipulated by adding more modules. Design and implementation of a wrench-based controller draws its inspiration from the realm of force-controlled legged robotics. To the best of our knowledge, this is the first application combining proven legged locomotion control approaches and proprioceptive actuators to CDPR systems to achieve interactive operations. Finally, validation of assumptions and functionality are experimentally shown. The system lifted an 27.2 kg (60 lb) payload and with the operator exerting an average of 14.5\% of the payload's weight to control its motion. This novel system offers significant advancements in the field of robotics, enabling safe, efficient, and adaptable manipulation of heavy payloads in diverse and dynamic environments.


\vspace{+2mm}
\section{Methodology}
\vspace{-2mm}
\begin{center}
\begin{figure*}[t]
    \centering
    \includegraphics[width=\textwidth]{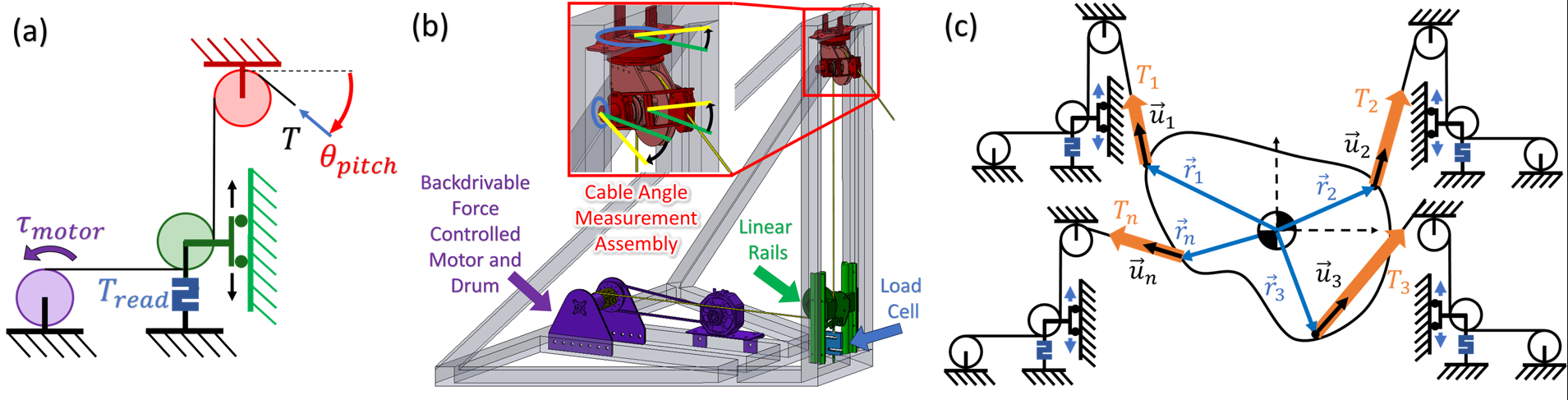}
\caption{(a) Simplified single module dynamics model representation of a Co3MaNDR Model. These are color coordinated with the (b) Robotic Module CAD model. The upper assembly measures tension output direction (yaw, pitch, and roll). The lower assembly moves up and down on rails ensuring that the measured cable tension is solely from the vertical component of the cable. (c) A generic number of modules (here represented as 1,2,3,n) manipulating a generic payload make up the system level dynamics diagram containing.}
\label{fig:module}
\vspace{-5mm}
\end{figure*}
\end{center}

\subsection{Scalable Force Based Controller}

Co3MaNDR's approach is built upon designing a system that enables assumptions which circumvent many physical interaction problems, thus simplifying control in multi-agent interactions. The core tenets are as follows: 1) minimize the effects of actuator dynamics, 2) treat each cable as an ideal force vector acting on the payload, and 3) optimize the vectors' magnitudes to achieve a desired outcome. This approach draws inspiration from the realm of force-controlled legged robotics~\cite{QP_1,QP_2}, bringing these proven control methods and hardware into CDPRs. By synergistically combining subfields, Co3MaNDR's controller deviates from the traditional cable length control and exclusively regulates forces between modules and the collective payload.

Adopting force control has several compelling advantages. First, it enables the monitoring and precise control of interaction forces, facilitating the sensing and recognition of external forces. This capability is instrumental in detecting and comprehending external forces resulting from collisions with unknown objects, people, or environmental entities at the system level. Constructing the controller to recognize and adapt to these external forces significantly enhances the safety of humans, other equipment, and the surrounding environment when operating in close proximity to Co3MaNDR.

Moreover, force sensing introduces an invaluable dimension for human-robot interaction. The system can detect and respond to applied forces with minimal resistance, allowing humans to physically guide the payload in a behavior referred to as force amplification. This human-in-the-loop approach leverages human planning capabilities, empowering operators with a more intuitive and responsive means of control. This has the potential to open the door to use Co3MaNDR in widespread applications.


By incorporating force control, Co3MaNDR eliminates the need for complex forward or inverse kinematics calculations, bypassing resource-intensive path planning. This streamlined approach reduces the system's computational overhead, enhancing real-time responsiveness and adaptability.
\vspace{+2mm}
\subsubsection{Implemented Controller}
The payload-centric force controller is designed to control the net wrench acting on the payload. This controller takes in the current system state and desired payload wrench and determines the correct tensions to be applied to the payload by each module. The desired wrench, $W_{des}$, can come from teleoperation commands, gravity compensation, or a desired trajectory or pose.
\vspace{-1mm}
\begin{equation}
\textbf{W}_{des} = [\textit{F}_{x}, \textit{F}_{y}, \textit{F}_{z}, \textit{M}_{x}, \textit{M}_{y}, \textit{M}_{z}]_{des}
\end{equation}

The net wrench applied is determined from the Jacobian ($\textbf{J}$), which is a function of all $u_i$ (the ith tension unit vectors) and $r_i$ (the ith vector reaching from the Center of Mass to the $i$th cable’s connection point). The Tension vector, $T$, is a column vector of all applied tension magnitudes. Figure \ref{fig:module} shows labeled entities in a system level diagram. 
\vspace{-2mm}
\begin{equation}
\textbf{W}_{applied} = \textbf{J} T = 
          \begin{bmatrix}
          u_1, \dots, u_n\\
          r_1 \times u_1, \dots, r_n \times u_n
          \end{bmatrix} 
          \begin{bmatrix}
           T_{1} \\
           \vdots \\
           T_{n}
         \end{bmatrix}
\end{equation}

Since cables cannot push and can only pull, most CDPR systems utilize more cables than the number of Degrees of Freedom (DoF) being controlled, creating an over-constrained system and expanding the solution set to be infinitely large. To navigate this complexity, this controller relies on quadratic programming (QP) optimization, which finds the decision vector (in this case the Tension vector, $T$) that minimizes the cost function within a set of linear constraints. Co3MaNDR’s cost function, shown in  Eq. \ref{eqn:cost_func}, is formulated to impose a weighted penalty for any deviation between the desired and achieved net payload wrench, while distributing tensions between minimum and maximum bounds, controlled by the matrix $\textbf{A}$ and vector $\textit{b}$.
\vspace*{-1mm}
\begin{mini}
      {T}{\smashoperator{\sum_{\tab \tab \tab i \in \{x,y,z,\theta_x,\theta_y,\theta_z\} }} \textbf{W}_i (F_{i,des}-F_{i})^2 + \textbf{W}_T \smashoperator{\sum_{\tab j \in [1 \dots n]}}T_j^2}{}{}
      \addConstraint{\textbf{A}T \leq b}{}{}
      \label{eqn:cost_func}
\end{mini}
\noindent where $F_{i,des}$ and $\textbf{W}_i$ are the desired force magnitudes and weights respectively for the Cartesian position in the ith directions. $\textit{W}_t$ represents the tensions weight, which is added to spread the forces between all cables and avoid over-utilizing any single module. Since this term of the optimization is not as important as controller accuracy, $\textit{W}_t$ should be at least an order of magnitude less than the Cartesian force weights. All weights were found experimentally. 

This cost function's analytic representation can be reorganized into a standard QP formulation as seen in Eq. \ref{eqn:QP_rep}. Here $\textit{x}$ is the decision vector (tensions), $\textbf{P}$ is a positive definite matrix, and $\textit{q}$, is a matrix containing the linear terms. $\textbf{P}$ and $\textit{q}$ are functions of $u_i$, $r_i$, and $W_{des}$. 
\vspace{-2mm}
\begin{mini}
      {T}{C(P,q) = \frac{1}{2} \textit{x}^T \textbf{P} \textit{x} + \textit{q}^T x}{}{}
      \addConstraint{\textbf{A}T \leq \textit{b},}{\textbf{ A}\in\mathbb{R}^{2n \times 2n},}{\textit{ b}\in\mathbb{R}^{2n \times 1}}
      \label{eqn:QP_rep}
\end{mini}

This formulation scales easily with changing the size of the decision vector and adding the additional force effects into the $\textbf{P}$ and $\textit{q}$. Consequently, the system level control effort scales by adding or removing modules, facilitating Co3MaNDR adapting to varying operational demands. 

\subsubsection{Assumptions and Requirements}
While the controller outlined in this paper demonstrates an elegant and effective approach to robotic control, it is based on several underlying assumptions which must be considered.  

First, the robot's actuators are assumed to function as pure torque sources, disregarding actuator dynamics, friction, damping, and inertial effects. This assumption is made viable through deliberate choice of backdrivable actuators, designed for compliant force control operations and operating in regimes that minimize these neglected effects. Second, the controller assumes modules are rigidly connected to the environment, precluding any shifting or tipping. Third, cables are assumed to be ideal, i.e., massless, frictionless, and with infinite stiffness. This assumption is well-founded when using stiff cables that are light in comparison to the payload, and tensions remain under plastic deformations limits. The neglected effects are orders of magnitude smaller than operational forces, which therefore dominate the system equation. Finally, the implementation ignores dynamics, assuming payload/modules do not experience dynamic or abrupt motions. 

\vspace{-2mm}
\subsection{Implementation}
\subsubsection{Module Design}
The robotic modules, seen in Fig. \ref{fig:module}, were engineered to support the assumptions in the controller, house all necessary components within a structurally robust, comparatively lightweight, and stable structure, and provide easy mounting to the operational environment. 

The motor output is attached by toothed belts to a drum on which the cable is wound (purple). This offers flexibility to adjust pulley ratios and tune motor attributes such as speed, backdrivability, and torque capabilities. 

The lower pulley assembly is mounted on linear rails (green) and firmly secured by a load cell (blue) to provide precise closed-loop force feedback. The height of the lower pulley can be adjusted to ensure that the cable runs perpendicular to the load cell's measurement axis and linear rails, which results in the load cells solely measuring cable tension. 

In the upper pulley assembly (red), three sets of encoders are incorporated to measure the cables' yaw, pulley roll, and measurement arm pitch, characterizing the applied tension unit vector and displacement.

\begin{table}[h]
\centering
\caption{High level Bill of Materials for system with four modules} 
\vspace{-3mm}
\begin{tabularx}{3.3in}{|X|X|c|}\hline
\textbf{Item} & \textbf{Use} & \textbf{Qty} \\
\hline
Laptop & Labview/control interface &  1 \\
\hline
Westwood Robotics PB02P & Main actuators &  4 \\
\hline
NI cRIO-9082 & System RT computer &  1 \\
\hline
NI 9205 Module& Analog input card&  1 \\
\hline
NI 9264 Module& Analog output card&  1 \\
\hline
NI 9403 Module& Digital IO card &  1 \\
\hline
NI 9871 Module& RS485 bus card &  1 \\
\hline
NI 9870 Module& RS232 bus card&  1 \\
\hline
NI 9381 Module& Multifunction I/O for cRIO &  1 \\
\hline
VectorNav VN-100 IMU & Estimate system pose &  5 \\
\hline
4S LiPo Batteries & Main power for the system &  8 \\
\hline
24V Kobalt Battery & Power sources for Load cells &  4 \\
\hline
Load Cells and Amplifiers & Closed loop force feedback &  4 \\
\hline
AEAT-6010-A06 Encoders & Measure angle state &  12 \\
\hline
Vicon Vero Motion Capture System & Closed loop position feedback - 8 cameras &  1 \\
\hline
\end{tabularx}
\label{tab:BoM}
\vspace{-6mm}
\end{table}

\subsubsection{Hardware}
The experimental system consists of four modules. Figure \ref{fig:flow_chart} shows a high-level flow diagram connecting components and Table \ref{tab:BoM} details them. Main control runs in real-time on an NI Compact Rio (cRIO) with expansion cards for analog and digital input/output, RS485, and RS232 communication busses. All communications between modules and the cRIO are over physical wires. Each module contains three 12-bit encoders, 1 IMU, 1 load cell/amplifier, and an actuator. 

To support ideal cable assumptions, cables were selected to have minimal elongation, stay below 60\% load capacity during operations, and be pre-stretched.

\subsubsection{System Level}
\begin{center}
\begin{figure}[h]
    \vspace{-4mm}
    \centering
    \includegraphics[width=0.485\textwidth ]{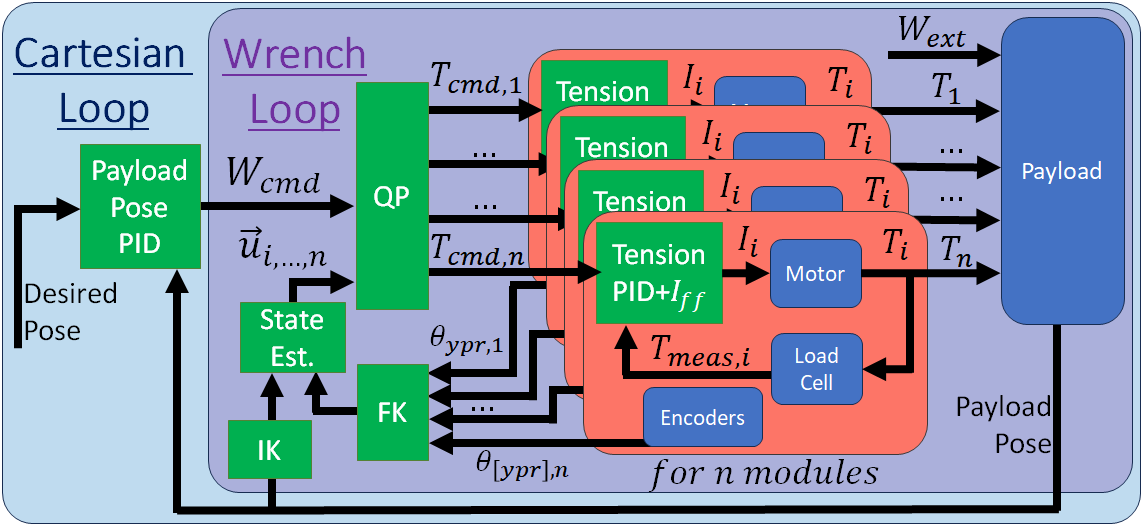}
    \vspace{-7mm}
\caption{System level control diagram for Co3MaNDR. Modules (red regions) contain individual $PID+I_{ff}$ loops, actuators, and sensors. The wrench based control loop (purple shaded region) shows the payload being controlled by any external wrench and applied tensions, converting from net wrench. The Cartesian control loop (light blue region) encapsulates the wrench loop and uses a pose $PID$ to regulate the desired wrench } 
\label{fig:flow_chart}
\end{figure}
\end{center}

The system’s control diagram when given force commands is shown within the purple zone in Fig. \ref{fig:flow_chart}. Payload wrench commands produce desired cable tensions. These are fed into a PID layer, minimizing steady state error and adding closed loop force feedback. External wrenches ($W_{ext}$) applied by operators guiding motion are introduced directly to the plant as an external disturbance. The blue zone in Fig. \ref{fig:flow_chart} houses the Cartesian control loop. Here, the PID pose controller encapsulates positional commands and closed-loop pose feedback from forward kinematics and a Vicon Vero Motion Capture system. 

\vspace{-8mm}

Rapid control rates have shown to be instrumental in force controlled robotics \cite{control_rate}. In the implementation, the proposed QP controller converges in 2ms on average, running at 500Hz with feedback from IMUs and Vicon data at 200Hz and encoders at 1kHz. The lower level PID reads load cells and sends control commands tensions at 1kHz.

\vspace{-2mm}
\section{Experiments}
A set of experiments was conducted to confirm the validity of the system and controller design assumptions, to verify that the prototyped system exhibits proposed requisite traits, and to demonstrate scaled component manipulation capabilities.
\subsection{Experimental Setup}

The experiments are limited to planar Cartesian position control with four modules seen in Fig. \ref{fig:entire_system}. The planar experimental setup consists of stacking one module on top of another and rigidly mounting them on top of concrete barriers, considered immovable at the operational force scale. Two identical setups were constructed facing each other creating a 4x2.5m planar workspace. The number of modules can be changed to match availability, lift heavier loads, or match the environment. Similarly, the number of controlled DoF can be adjusted to achieve desired manipulation effects.
\vspace{-12mm}
\begin{center}
\begin{figure}[t]
    \centering
    \includegraphics[scale=0.55]{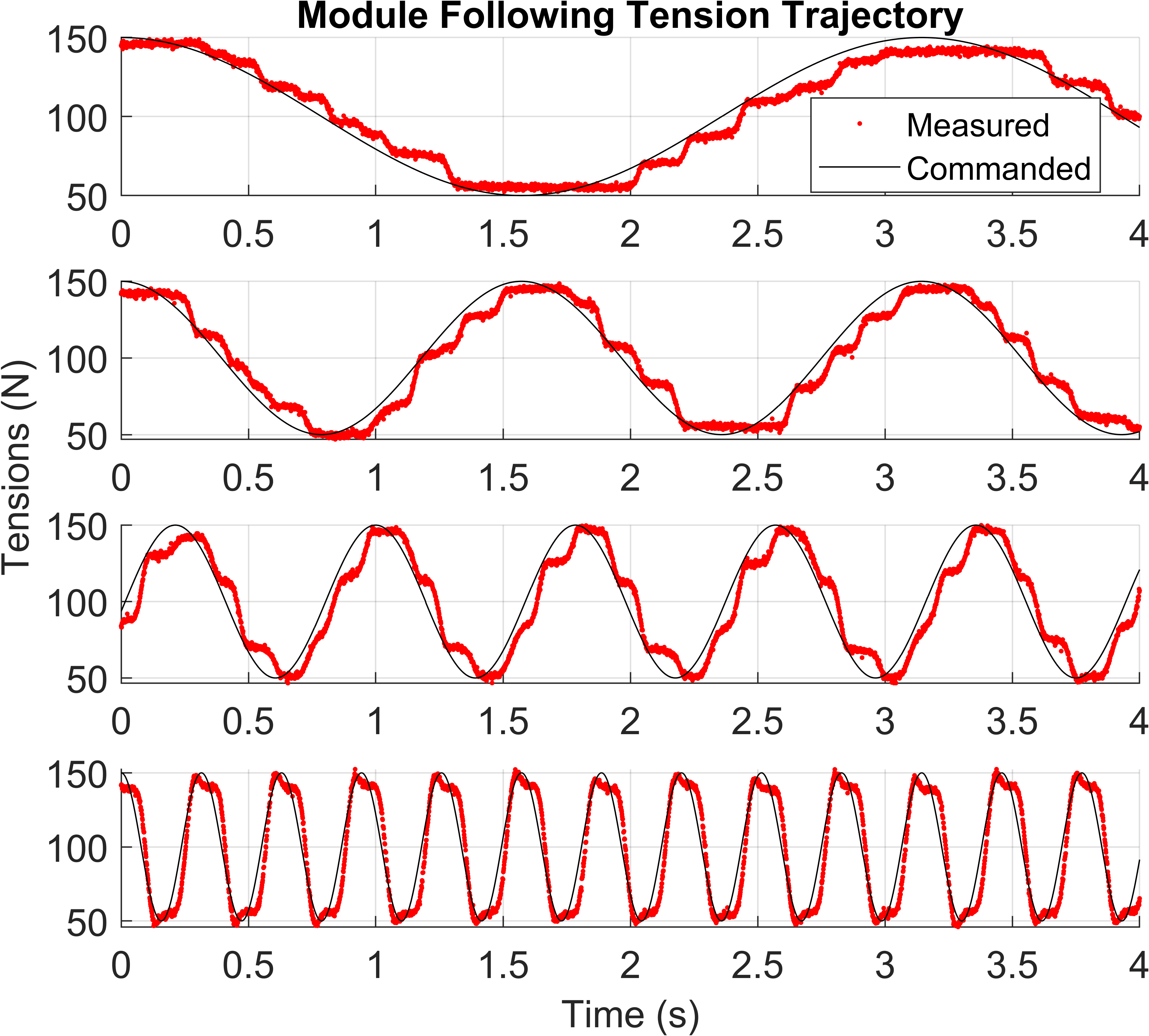}
    \vspace{-4mm}
    \caption{Co3MaNDR System following moving tension reference trajectory at several frequencies. Despite minimal lag and exhibited stiction stair-stepping effects, the desired force trajectory is followed closely.} 
\label{fig:moving_traj}
\vspace{-3mm}
\end{figure}
\end{center}
\subsection{Module Characterization}
The proposed control scheme is predicated on the assumption of ideal force sources. The first set of experiments characterizes module performance and ascertains to what degree they may be represented as a pure tension source. 

In the initial phase of this experiment, we examine modules exerting force by rigidly connecting the module and cable to the environment. Starting from the minimum allowed torque (30N), we command 4 steps cable tension in 50N increments while recording the actual forces. The subsequent phase investigates the modules' ability to absorb forces, focusing on their level of backdrivability. While commanding a constant tension - 0N to 200N in 50N increments - error is introducing by manually displacing the cable.

\subsection{Scaling and Modularity}
The second experiment was designed to exhibit control scaling, physical reconfiguration, and modularity attributes while showing teleoperation and trajectory following abilities. This experiment was subdivided into two parts. In each, the system moved the same payload in the same 1m square trajectory. Experiments only differed by using either the upper two or all four modules. The performance of each module and system as a whole was compared. 

\subsection{Manipulation and Force Amplification}
The final experiment was designed to show system compliance, force control, safe interactions, and effort amplification. It consisted of a human operator pulling a load cell attached to the 27.2kg (60 lb) payload in both positive and negative $X$ and $Z$ directions while Co3MaNDR compensated for the payload's weight. The payload motion, system effort, and manually applied wrench were recorded and compared to the payload's weight. 
\vspace{-8mm}

\begin{center}
\begin{figure}[t]
    \centering
    \includegraphics[width=0.46\textwidth]{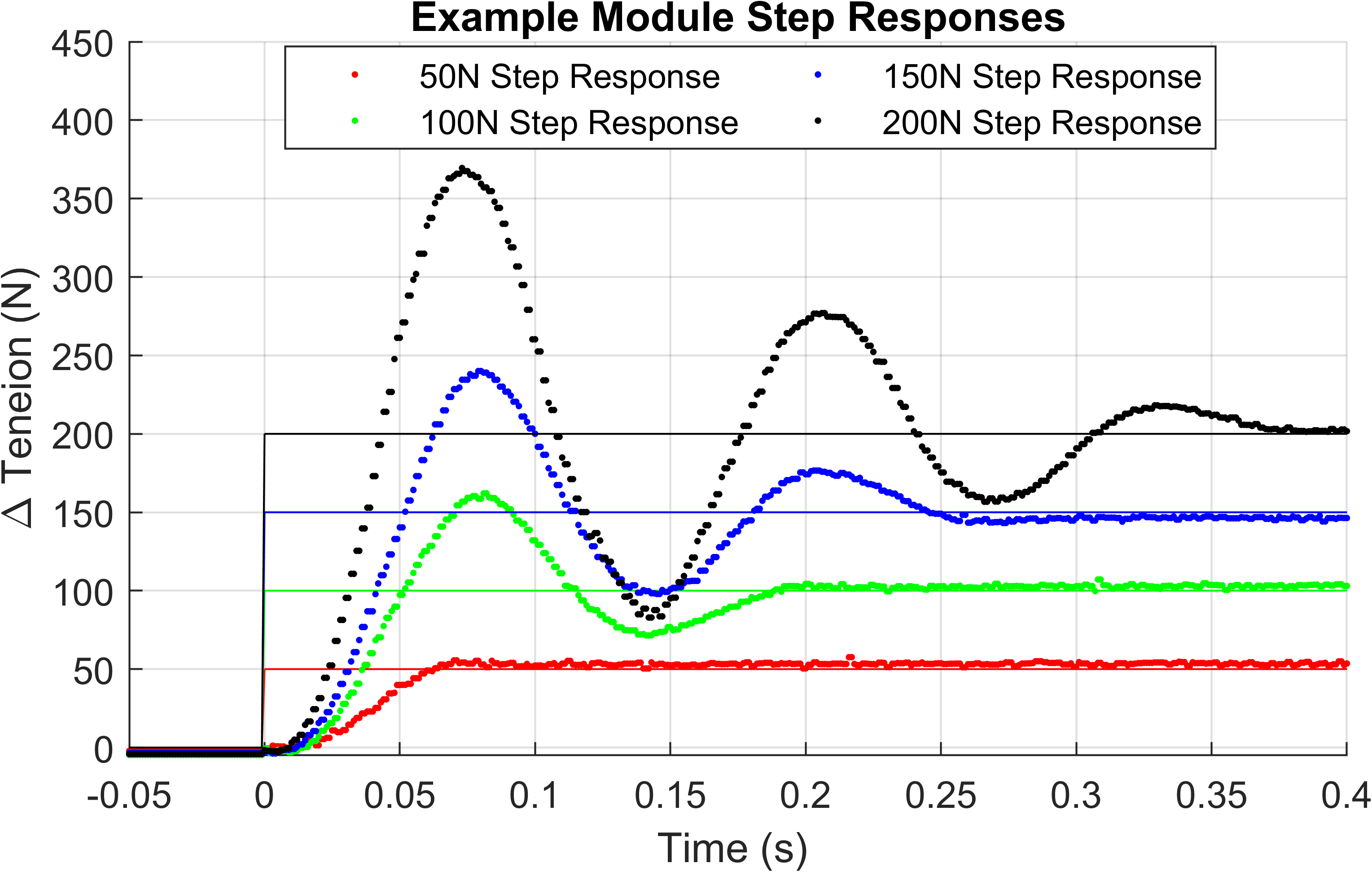}
    \vspace{-4mm}
\caption{Module characterization for step responses of several commanded tension magnitudes. Each experiment showed rise times < 0.7ms and settling times between 0.75-3.8 ms. Larger amplitude steps had large overshoots. In use, commanded tensions will not have large jumps, so performance in 50N step response is considered fitting and sufficient for operations.} 
\label{fig:steps}
\vspace{-4mm}
\end{figure}
\end{center}


\section{Results and Discussion}
In this section, we present and discuss experimental results, highlighting Co3MaNDR's performance and its implications.  Beyond demonstrating system capabilities, the analysis also offers insights into technology improvement areas, future research directions, and potential applications.

\vspace{-8mm}
\begin{center}
\begin{figure*}[t]
    \centering
    \includegraphics[scale=0.5]{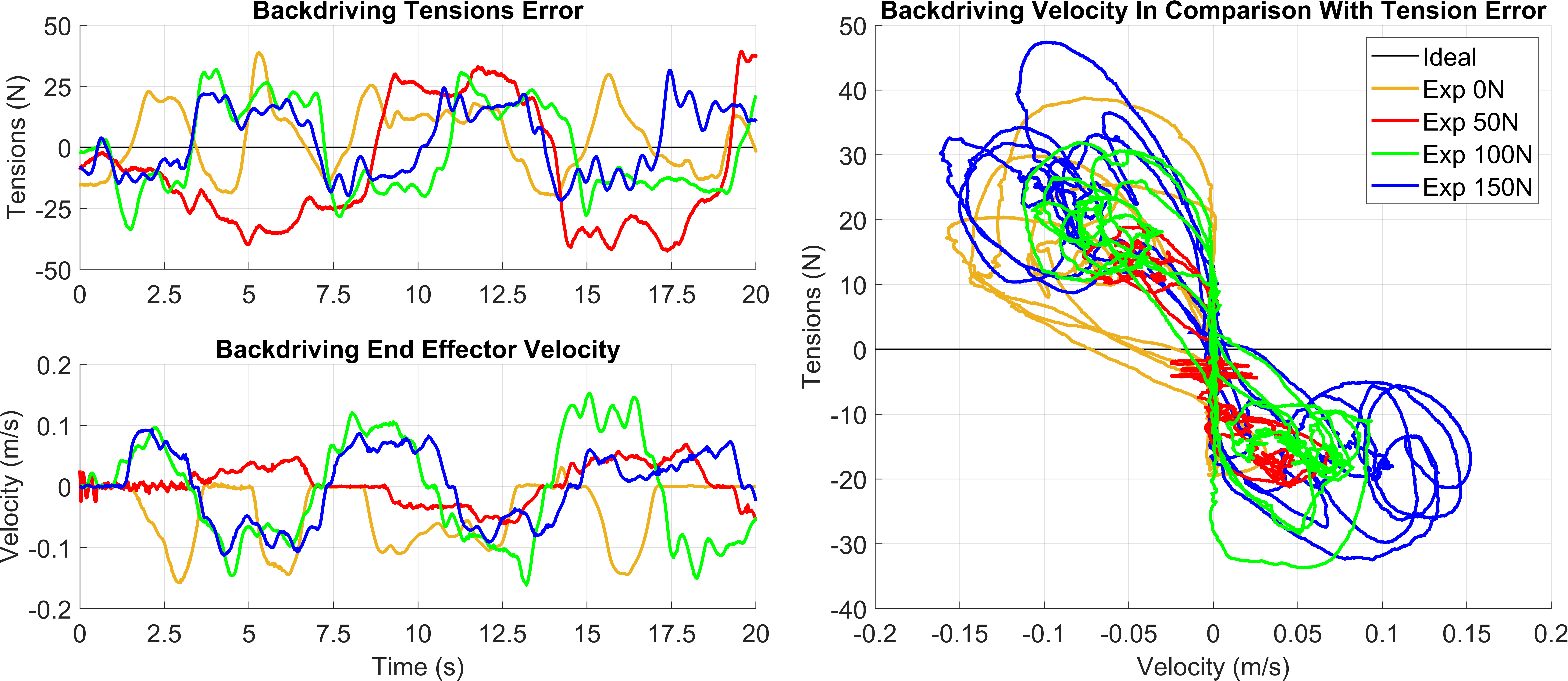}
    \vspace{-3mm}

\caption{Module characterization for back-driving during constant commanded tension. Top left shows that the tension error stays within the same range despite tension magnitude. Bottom left show the velocity of the cable, exhibiting stiction behaviors. Right graph shows the tension error vs velocity relationship, finding a negative correlations between the two. Stiction is also depicted at $\pm$ 10N on the Y (Tension) axis}
\vspace{-6mm}
\label{fig:backdrive}
\end{figure*}
\end{center}
\subsection{Module Force Control Performance}
The module design targets friction minimization, but it cannot be completely eliminated. Friction most clearly expressed itself through non-uniform end-effector task-space inertia. Any backdriving displacing a motor must overcome its reflected inertia and motor damping. Moving the cable output along the tension vector results in the maximum motor backdriving while moving perpendicular has minimal backdriving. In these experiments, the tensions vectors had much larger X than Z components. Therefore, motion in directions opposite of the tension vectors ($\pm X$) backdrove the system’s total set of motors significantly further than motions that travelled orthogonal to them ($\pm Z$). The resulting directionally dependant apparent inertia made moving the payload along the X axis feel heavier; similarly, motions in the Z direction felt lighter and were easily achieved manually. Testing without the closed loop force feedback confirmed and accentuated these results.

\subsubsection{Force Control: Exerting Forces}
To verify modules as pure torque output sources, Fig. \ref{fig:moving_traj} compares desired and measured tension from a module commanded sinusoidal trajectories. Stiction is seen in the lagging stair-stepping effect, smaller adjustments in tension and motor motion. The magnitude of the error is proportionally small compared to control effort, so it viewed as disturbance. 


Figure \ref{fig:steps} shows a module's step responses at four magnitudes. All rise times are universally within 75 ms. Average human response times are 250-400ms through sensing, deciding, planning, and beginning the action \cite{neuro}. Since Co3MaNDR's response times are faster, the system is sufficiently fast to mimic human manual assembly operations. The overshoot seen in the larger steps is a product of the feed forward current and PID controller compounding efforts at the initial command. While overshoot can cause large oscillations with large steps, the 50N step has virtually no overshoot and settling time under 100ms. Tension trajectories during several experiments, seen in Fig \ref{fig:ten_mag_compare}, show that changes in operational commanded tension are significantly less abrupt than a 50N step. As Co3MaNDR is used in relatively slow (human assembly speed) operations, operational performance is skewed toward the lower magnitude steps responses and overshoot becomes a non-issue.

\subsubsection{Backdriving: Absorbing Forces}
Figure \ref{fig:backdrive} shows measured tension error and velocity during backdriving experiments. Stiction again presents itself in the signals on the right graph. Here, signals run along the vertical zero-speed axis between approximately $\pm 10N$. Above that, the signals sharply break away, forming two lobes in the second and fourth quadrants. Zero commanded tension tests skew away from this trend due to the lower pulley assembly's weight not being held up by a tensioned cable. 

Interestingly, tension error magnitude does not increase with higher exerted forces, showing disturbances do not scale with control effort. Results indicate that proprioceptive force controlled CDPRs can be scaled up to larger tensions without greatly increasing force error. The fact that control effort error did not scale with effort magnitude supports the conclusion that the error is not caused by control failures, but instead by unavoidable backdriving effects. 

These experiments confirm that the modules successfully achieve force control and act as a pure torque source during motions within a human manipulation speed regime.

\begin{center}
\begin{figure}[h]
    \centering
    \includegraphics[width=0.485\textwidth]{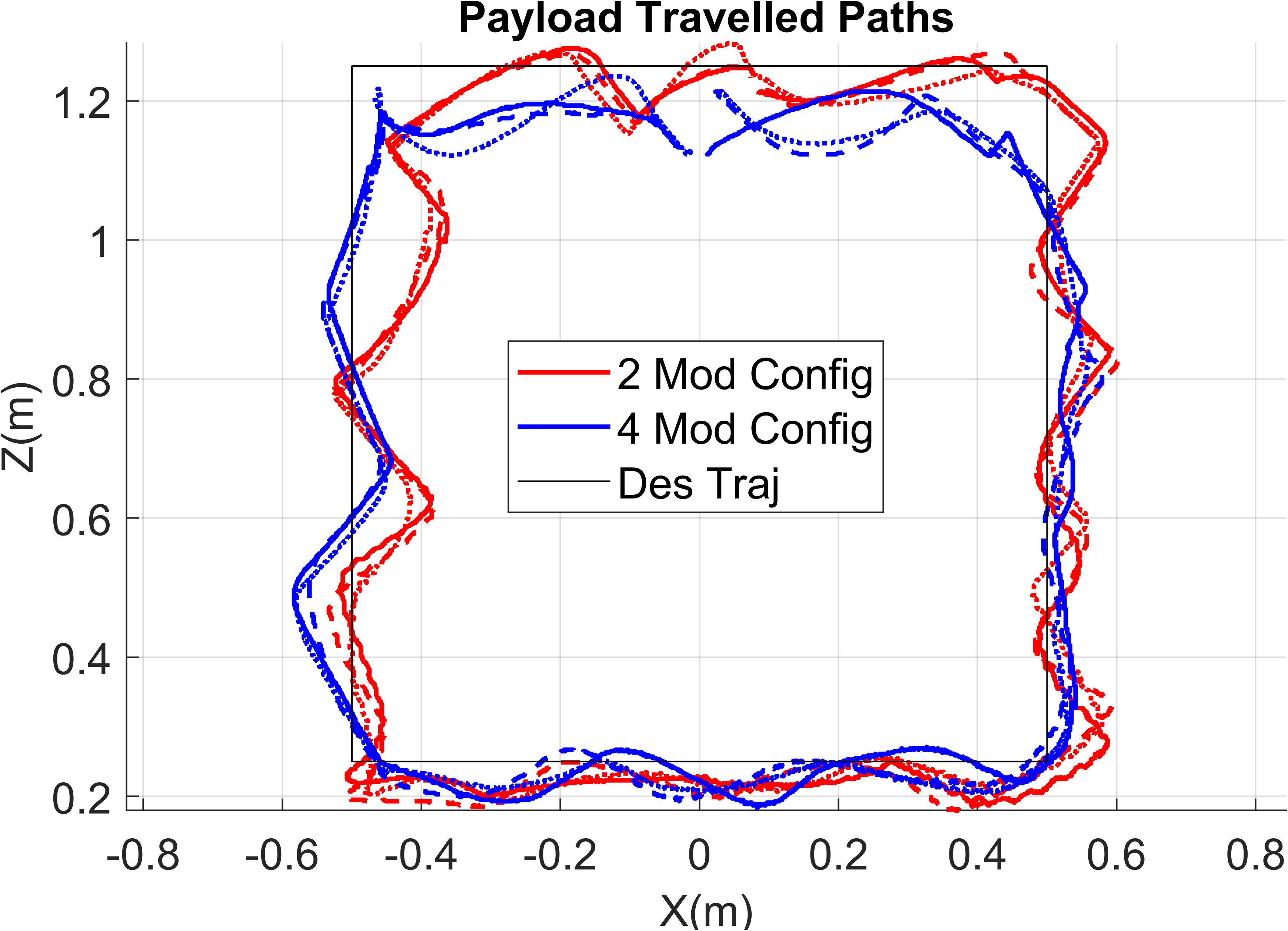}
    \vspace{-7.00mm}
\caption{Paths comparing the end-effector's motion during experiments with two and four module configurations. Motions are repeatable and comparable. Stiction is seen in zig-zagging in payload motion. } 
\label{fig:motion_compare}
\vspace{-5.00mm}
\end{figure}
\end{center}

\vspace{-8.00mm}

\subsection{Scaling and Modularity}
\begin{center}
\begin{figure}[t]
    \centering
    \includegraphics[width=0.485\textwidth]{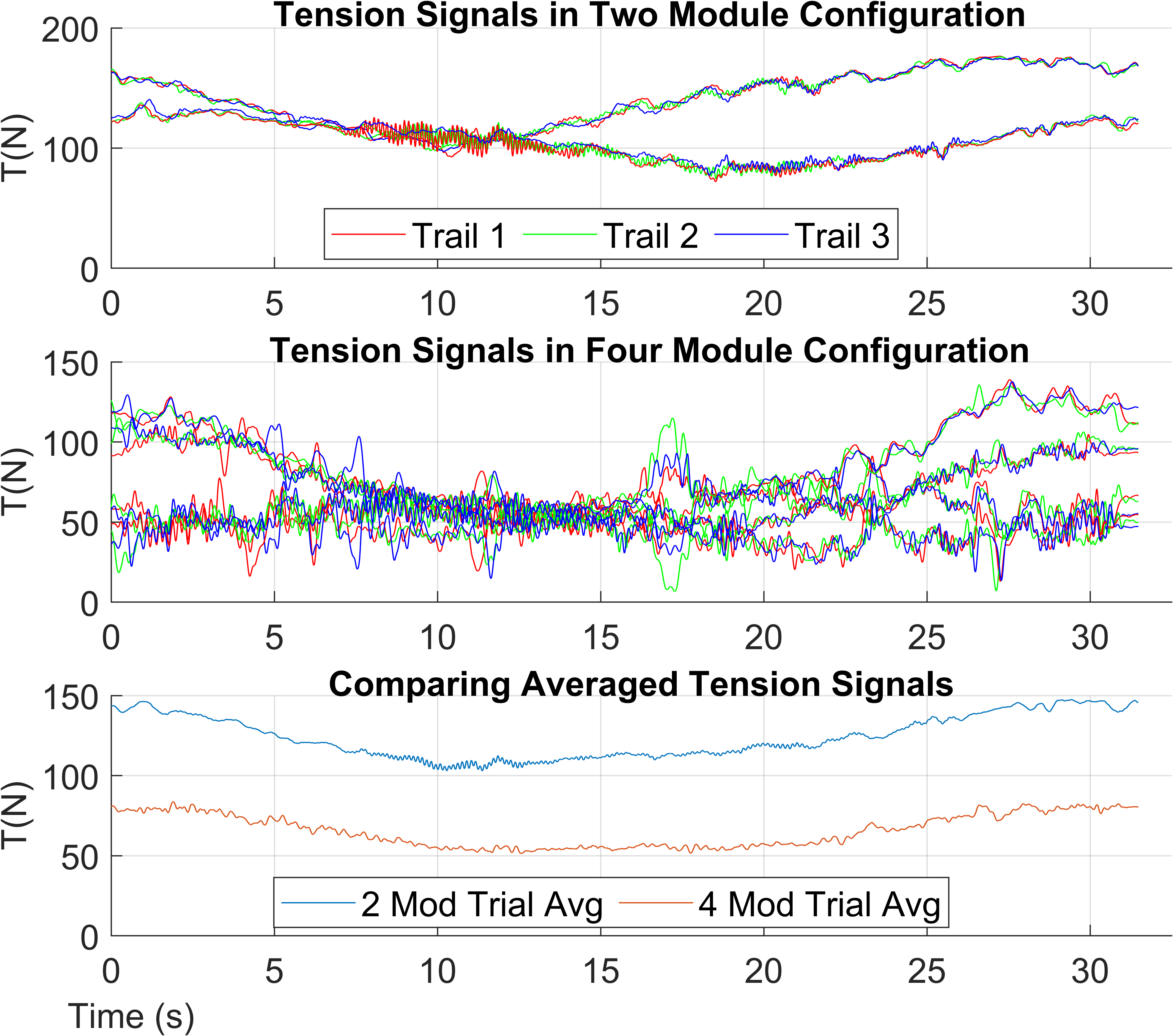}
    \vspace{-7.00mm}
\caption{Each module's tension trajectories during two module experiments (top), four module experiments (middle), and their respective averages (bottom). Average tension in four module experiments is consistently half that of two module experiments showing scalability in control and capabilities, reconfiguability, and modalarity.} 
\label{fig:ten_mag_compare}
\end{figure}
\end{center}

Several trials' commanded and executed trajectories, consisting of either two or four modules, are shown in Fig. \ref{fig:motion_compare}. Both configurations achieved the desired macroscopic payload trajectory, with a max error of only 13.7 cm in a 4 x 2 m workspace. Shifting focus from positions to forces can incur higher errors. Intuitively, this error originates from stiction; once stationary, the payload builds up error until control effort overcomes static friction. This effect is seen in the trajectories as a series of linear motions connected by directional changes and temporal pauses. Figure \ref{fig:ten_mag_compare} shows the tension signals in both configurations in their comparable operations. The average tension in trails using four modules was half the tension of trails using only using two. Additionally, the ratio of maximum tension achieved between the four to two module modes is $\sim$75\%.

Intuitively, the four module configuration spreads out the torque requirements to more modules, so used proportionally less of each module’s maximum torque capabilities. This not only is safer for the system, but it implies that more modules increase system lift capacity, which confirms scalable control on a physically reconfigurable modular system.

\subsection{Manipulation and Force Amplification}
\begin{center}
\begin{figure}[t]
    \centering
    \includegraphics[scale=0.5]{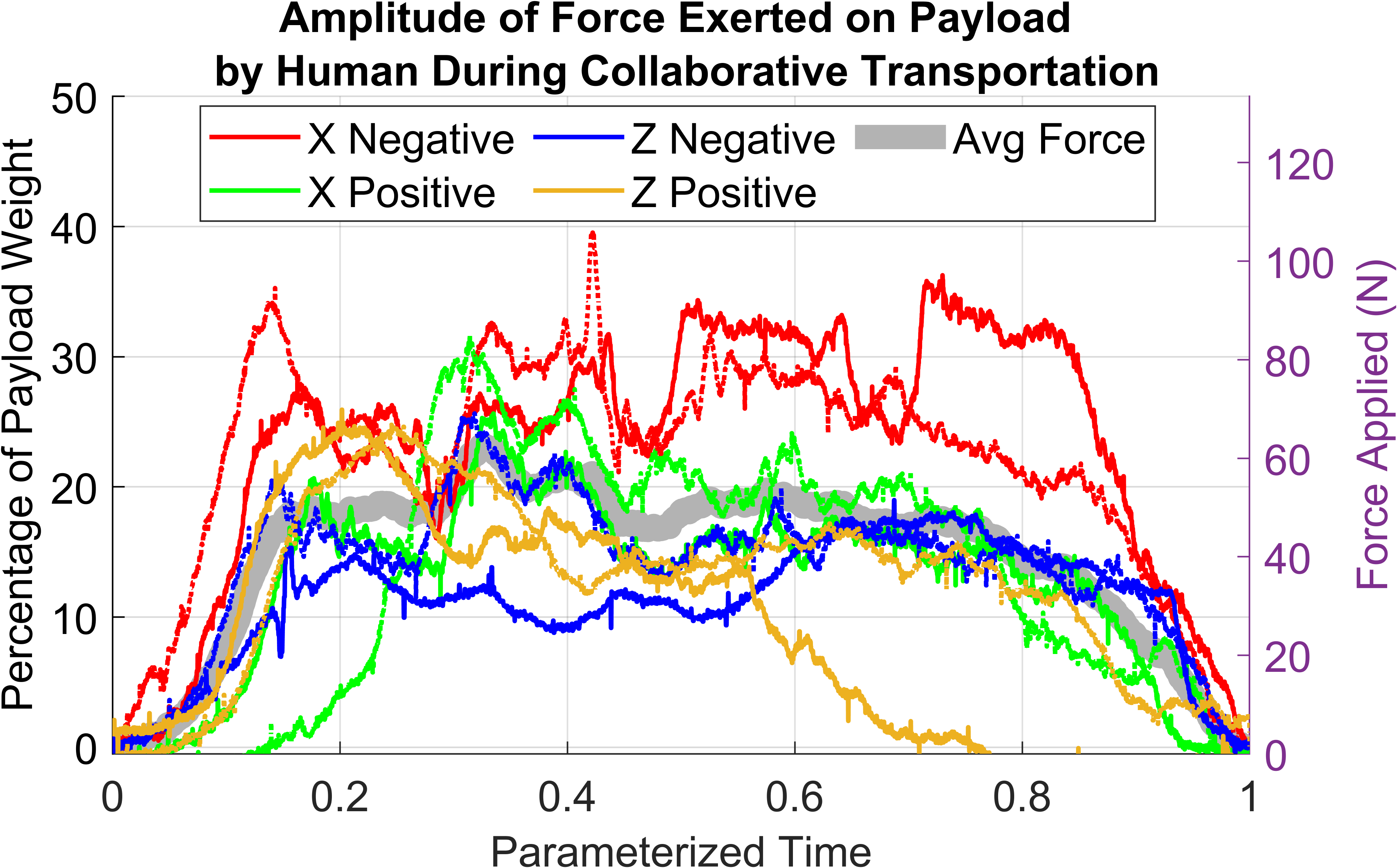}
    \vspace{-3mm}
\caption{Forces human exerted on a payload supported by Co3MaNDR during force amplification and manual manipulation. Maximum and average effort is around 39.4\% and 14.5\% of the payloads weight respectively.} 
\label{fig:measured_tension}
    \vspace{-7mm}
\end{figure}
\end{center}

Figure \ref{fig:measured_tension} shows both the payload weight and the measured operator-to-payload applied forces during interactive manipulation. The operator exerts an estimated maximum force of approximately 105.2 N and average approximately 38.7 N, only 39.4\% and 14.5\% of the payload’s weight respectively. Overcompensating for friction leads to instability~\cite{fric_comp}. Similarly, amplifying applied forces to eliminate apparent inertia often leads to instability. Therefore, payload inertia is treated as an irremovable constant. Co3MaNDR demonstrates sufficiently high transparency and force control to stably allow collaborative and unplanned interaction without additional external sensing. This successfully reduces the force required for a human to manipulate a heavy object safely.




\vspace{-4mm}

\vspace{-13mm}
\subsection{Limitations}
\subsubsection{Proportions and Scale}
The most obvious limitation of the system is the reduced lifting capacity compared to motors with traditionally high gear ratios. The prototype was designed and components chosen based off simulated values performing bridging operations scaled down (for practicality) to move a 30kg payload throughout the workspace. Since the tension error magnitude was found to not scale with control effort, using more massive payloads make frictional and backdriving effects proportionally smaller. However, the assumptions limit the extent of proportional scaling. Larger, thicker, and heavier cable required for significantly higher loads introduce cable sag, bending limitations, and elongation effects, which are more significant at larger scales. 

\subsubsection{Actuator Balance}
SEAs, QDDs, hydraulic, or other types of actuators can be implemented for a full-scale design. The less the motors act as an ideal torque source, the less effectiveness the controller, and therefore the more reduced system capabilities. The less ideal the actuators, the greater the antagonistic forces between modules and that operator's manual guidance must overcome. Increasing the transparency and torque capacity of the system can grant significantly larger force multiplication, allowing the same person to manipulate a comparatively more massive payload in the same manner. Selecting motor properties, for instance backdrivablity, mechanical bandwidth, friction, and sustained and burst torque capabilities, to best replicate an ideal torque source in the intended application is a balancing act.

\subsubsection{Controlling Pose}
Manipulating objects' spatial position is useful. However, bridging, construction, and other applications require controlling the position and orientation of the payload. The controller is designed to expand control to additional DoF; simulations confirmed the validity of this approach. However, controlling more DoF also places limits on the possible number of modules, configurations, and the wrench feasible workspace~\cite{wrench_feasible_workspace}. 

\subsubsection{Operational Mobility}
These experiments utilize static modules, limiting their field utility. Transporting a massive payload using this approach would necessitate repeatedly setting up a workspace, moving the payload, and finally moving the workspace to an advanced position. While mobile CDPRs do exist~\cite{FASTKIT,mcdpr}, they generally require specialized platforms moving in coordination in ideal and well-known environments. Adding long distance field transportation capabilities to this technology increases potential mission capabilities in civil, industrial, and military scenarios. 


\section{Conclusion}
To address the unique combination of challenges and limitations inherent in current field bridging, this research proposed solutions must have the following attributes: modular, physically reconfigurable, scalable control, force controlled, and compliant. These attributes shaped the Co3MaNDR system’s design and implementation.


Co3MaNDR was verified with a set of experiments. In during motions matching human speeds, Co3MaNDR was shown to achieve force control sufficiently quickly and accurately to collaborative amplify externally applied forces and comply with unplanned interaction. These capabilities reduced the  force required for a single operator to physically manipulate a cumbersome payload with an average of 14.5\% of the payload’s weight. Results show that Co3MaNDR's lifting capabilities scale with the number of employed modules, showing that the physical system and controls can accommodate a large range of use-cases. Further, experiments indicate that cable tension’s error did not scale with tension magnitude; this indicates that scaling up the control effort proportionally reduces effects of stiction and friction. 

This technology has the potential to remove the need for heavy equipment in field construction operations, reducing logistical burdens, increasing environmental suitability, and vastly increasing transportation and construction speeds. Additionally, the system's proprioceptive nature can potentially improve operator safety during assembly and transportation operations. Adoption of this approach may reduce manual requisite manual labor and improve personnel safety. 

Co3MaNDR's future research will explore additional controlled payload DoFs, the mobility and coordination of modules, incorporated vision systems, and actuation approaches. These advances will expand the applicability and utility of mobile CDPRs. 
\vspace{-4mm}
\bibliographystyle{ieeetr}
\bibliography{main}

\end{document}